\documentclass[article]{jss}

\usepackage{subfigure}
\usepackage{makeidx}         
\usepackage{amsmath}
\usepackage{amsfonts}
\usepackage{amssymb}

\usepackage{graphicx}        
\usepackage{multicol}        
\usepackage{algorithmic}
\usepackage{algorithm}
\usepackage{epstopdf}

\def\R{\mathbb{R}}

\def\F{\mathcal{F}}
\def\G{\mathcal{G}}

\def\mle{\mathrm{EM}}
\def\cec{\mathrm{CEC}}

\def\x{\mathrm{x}}

\def\tr{\mathrm{tr}}
\def\cl{\mathrm{cl}}
\def\m{\mathrm{m}}
\def\I{\mathrm{I}}

\def\y{Y}

\usepackage{lscape}
\usepackage{rotating}

\usepackage{lscape}
\usepackage{rotating}

\usepackage{listings}
\usepackage{color}
\usepackage{array}

\author{Jacek Tabor\\Jagiellonian University \And 
		   Konrad Kamieniecki\\ Jagiellonian University \And 
           Przemys\l{}aw Spurek\\Jagiellonian University \AND 
           Krzysztof Misztal\\AGH University of Science and Technology \And 
           Marek \'Smieja\\Jagiellonian University
          }
\title{Introduction to Cross-Entropy Clustering\\
The \proglang{R} Package \pkg{CEC}}

\Plainauthor{Jacek Tabor \And Konrad Kamieniecki, Przemys\l{}aw Spurek \And Krzysztof Misztal \And
Marek \'Smieja}
\Plaintitle{Introduction to Cross-Entropy Clustering} 
\Shorttitle{Cross-Entropy Clustering} 

\Abstract{
The \proglang{R} Package \pkg{CEC} \cite{CEC} performs clustering based on the cross--entropy clustering (CEC) method, which was recently developed with the use of information theory. 
The main advantage of CEC is that it combines the speed and
simplicity of $k$-means with the ability to use various Gaussian mixture models and reduce unnecessary clusters.
In this work we present a practical tutorial to CEC based on the \proglang{R} Package \pkg{CEC}.
Functions are provided to encompass the whole process of clustering.   
}
\Plainkeywords{clustering, Gaussian mixture models, density estimation} 

\Address{
  J. Tabor, P. Spurek, K. Kamieniecki, M. \'Smieja\\
 Jagiellonian University\\
 Faculty of Mathematics and Computer Science\\
 \L{}ojasiewicza 6, 30-348 Krak\'ow, Poland\\
  E-mail: \email{tabor@ii.uj.edu.pl}, \email{przemyslaw.spurek@ii.uj.edu.pl},\\\email{konrad.kamieniecki@uj.edu.pl}, \email{marek.smieja@ii.uj.edu.pl}
  
K. Misztal\\
AGH University of Science and Technology\\
Faculty of Physics and Applied Computer Science\\
al. A. Mickiewicza 30, PL-30059 Krak\'ow, Poland\\
  E-mail: \email{Krzysztof.Misztal@fis.agh.edu.pl}
}

\begin{document}

\section{Introduction}

Clustering plays a basic role in many parts
of data engineering, machine learning, pattern recognition and image analysis, see \cite{Clu, Dubes, jain1999, jain2010,xu2009clustering}. Thus, it is not surprising that numerous clustering methods were implemented as an \proglang{R} Package e.g.
\pkg{mclust} (\cite{fraley1999mclust}), \pkg{pdfCluster} (\cite{Azzalini2013}), \pkg{mixtools} (\cite{Benaglia2009}), \pkg{clues} (\cite{Chang2010}), \pkg{HDclassif} (\cite{Berge2011}), \pkg{ClustOfVar} (\cite{Chavent2012}), etc.

\begin{figure}[!t] \centering
	\subfigure[$k$-means with $k=3$ initial number of clusters.]{\label{fig:mouse_1}\includegraphics[width=0.3\textwidth]{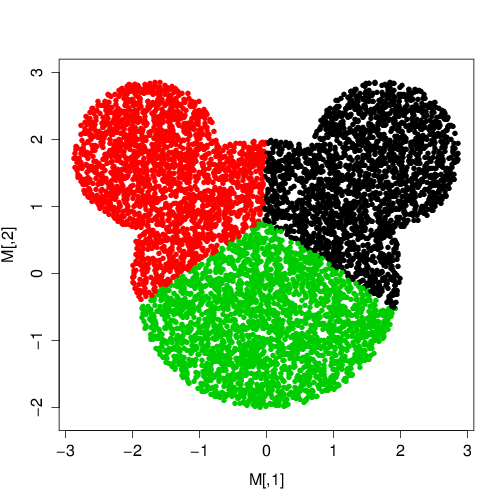}}\qquad
	\subfigure[$k$-means with $k=10$ initial number of clusters.]{\label{fig:mouse_2}\includegraphics[width=0.3\textwidth]{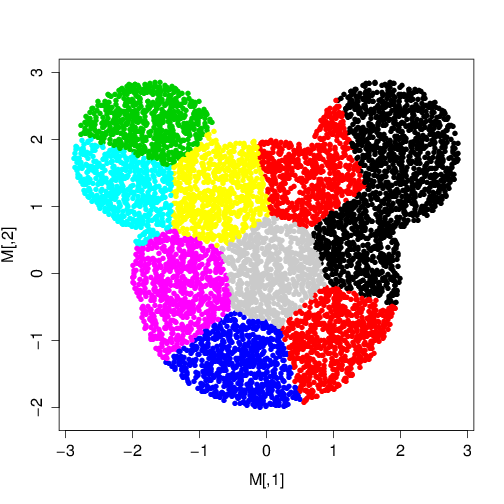}} \\		
		\subfigure[Spherical Mclust with $k=3$ initial number of clusters.]{\label{fig:mouse_3}\includegraphics[width=0.3\textwidth]{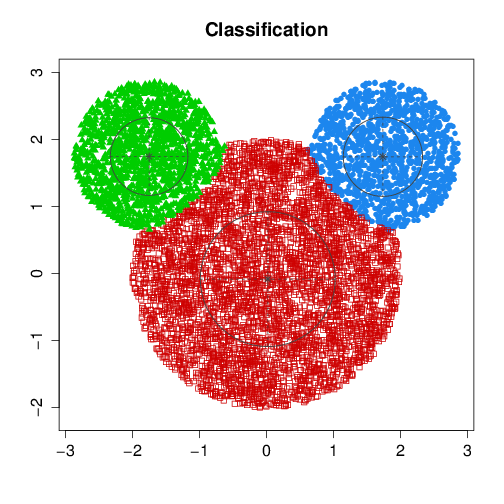}}\qquad
		\subfigure[Spherical Mclust with $k=10$ initial number of clusters.]{\label{fig:mouse_4}\includegraphics[width=0.3\textwidth]{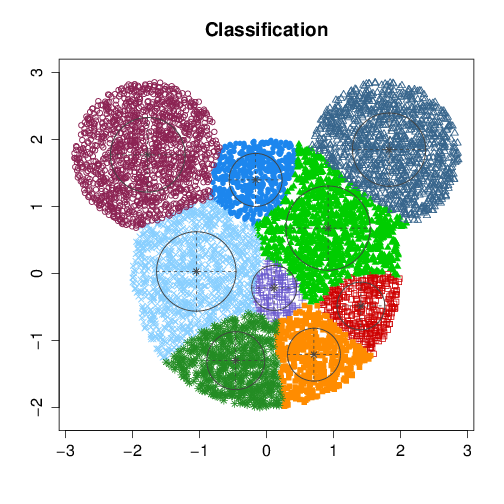}}\\
			\subfigure[Spherical CEC with $k=10$ initial number of clusters, which was reduced to $k=3$.]{\label{fig:mouse_5}\includegraphics[width=0.3\textwidth]{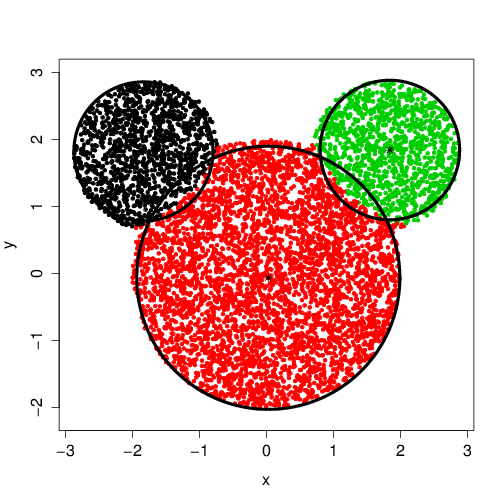}}\qquad
	\subfigure[General Gaussian CEC with $k=10$ initial number of clusters, which was reduced to $k=6$.]{\label{fig:mouse_6}\includegraphics[width=0.3\textwidth]{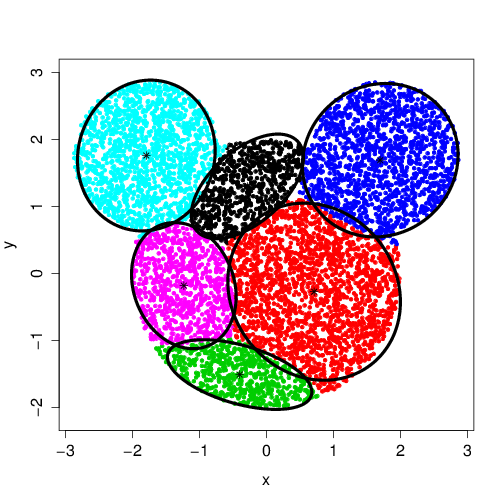}}
	\caption{ Clustering of the uniform density on a mouse-like set by various types of algorithms.}
	\label{fig:mouse} 
\end{figure}

Several of the most popular clustering methods are based on the $k$-means  approach, see \cite{bock2007, bock2008}. Although $k$-means is easily scalable, 
it has the tendency to divide the data into
spherically shaped clusters of similar sizes. Consequently, it is not affine invariant and does not deal well with clusters of various sizes. This causes the so-called mouse-effect, see Fig.~\ref{fig:mouse_1}. Moreover, it does not change dynamically number of clusters, see Fig.~\ref{fig:mouse_2}, and therefore in order to efficiently apply $k$-means, usually data preprocessing (like whitening) needs to be applied and additional tools like gap statistics developed by \cite{gap,mirkin2011choosing} to choose the right number of groups have to be used. 

Another group of clustering methods is based on density estimation techniques which use Expectation Maximization (EM) method (\cite{EM2, EM3}). Probably the Gaussian Mixture Model (GMM) is the most popular, see
\cite{mclachlan2007algorithm,mclachlan2004finite}.
It is hard to overestimate the role of GMM and its generalizations in computer science  (\cite{mclachlan2007algorithm,
mclachlan2004finite,
Dubes
}) in particular in object detection (\cite{huang1998extensions,figueiredo2002unsupervised}, object tracking (\cite{xiong2002improved}), learning and modeling (\cite{samuelsson2004waveform}), feature selection (\cite{valente2004variational}), classification (\cite{povinelli2004time}) or statistic background subtraction.

The relation between the above two methods is well described by \cite{estivill2000fast}:
"[...] The weaknesses of $k$-means results in poor quality clustering, and
thus, more statistically sophisticated alternatives have been proposed.
[...] While these alternatives offer more
statistical accuracy, robustness and less bias, they trade this for substantially more computational requirements and more detailed prior knowledge, see \cite{massa1999new}."

The Cross--Entropy Clustering (CEC) approach proposed by \cite{tabor2014cross} joins the clustering advantages of $k$-means and EM. It occurs that CEC inherits the speed and scalability of $k$-means, while overcoming the ability of EM to use mixture models. In particular, contrary
to GMM, new models can easily be added without the need
for complicated optimization (Section \ref{cha:mix}).
Consequently, this allows the use of CEC as an elliptic pattern recognition tool (\cite{Ta-Mi,Sp-Ta}).
The motivation of CEC comes from the observation that in the case of coding
it is often profitable to use various compression algorithms
specialized in different data types. 
The idea was based on the classical Shannon Entropy Theory, see \cite{Co-Th, Ka, Sh} 
and the Minimum Description Length Principle (\cite{MDLP, Gr}). Similar approach, which uses MDLP for image
segmentation, was given by \cite{Yi_Ma,Yi_Ma2}.
A~close approach from the Bayesian perspective
can also be found in the works of
 \cite{kulis2012revisiting,kurihara2009bayesian,korzen2013logistic}.

\begin{figure}[!t]
\centering
\includegraphics[width=0.15\textwidth]{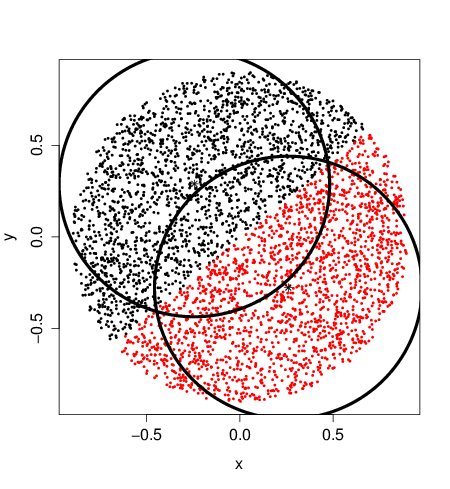}
\includegraphics[width=0.15\textwidth]{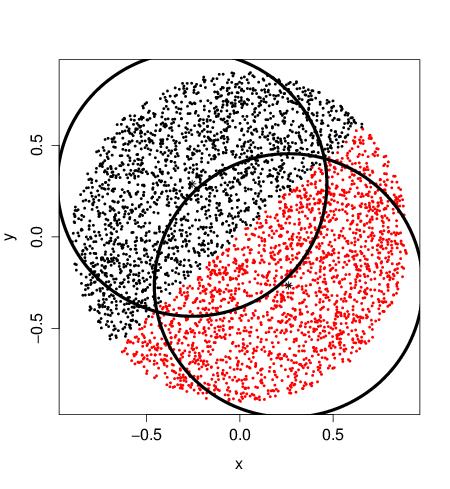}
\includegraphics[width=0.15\textwidth]{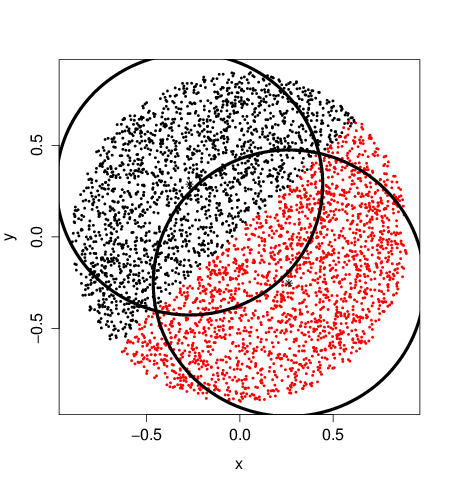}
\includegraphics[width=0.15\textwidth]{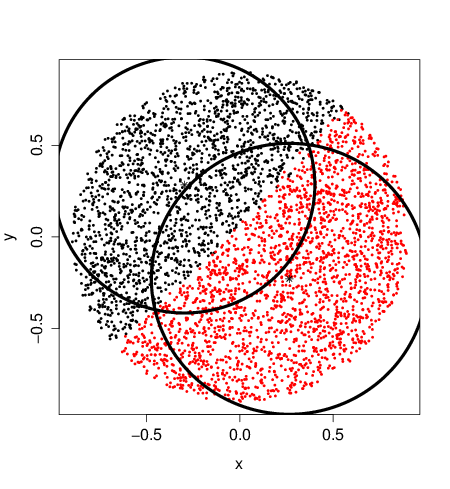}
\includegraphics[width=0.15\textwidth]{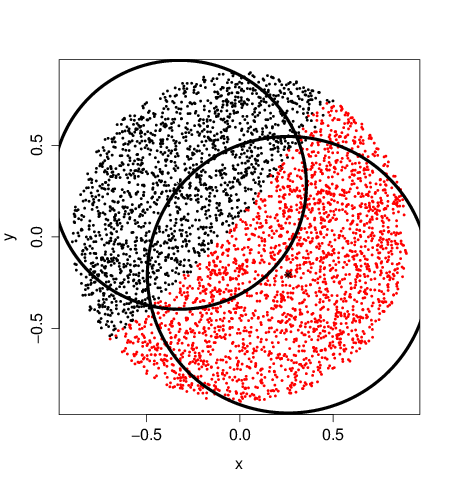}
\includegraphics[width=0.15\textwidth]{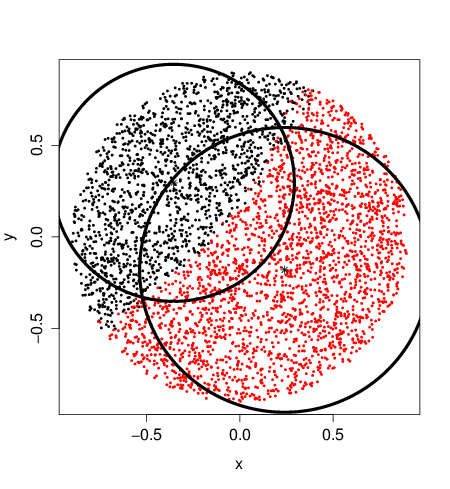}
\includegraphics[width=0.15\textwidth]{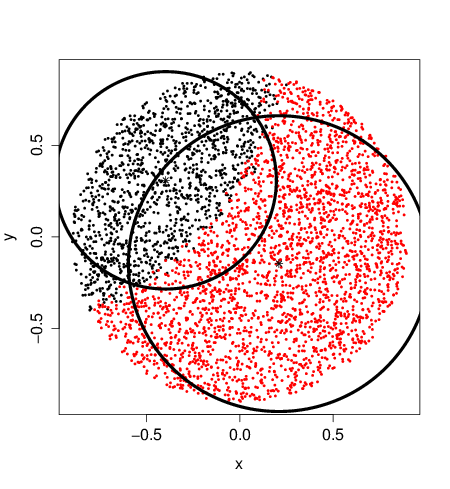}
\includegraphics[width=0.15\textwidth]{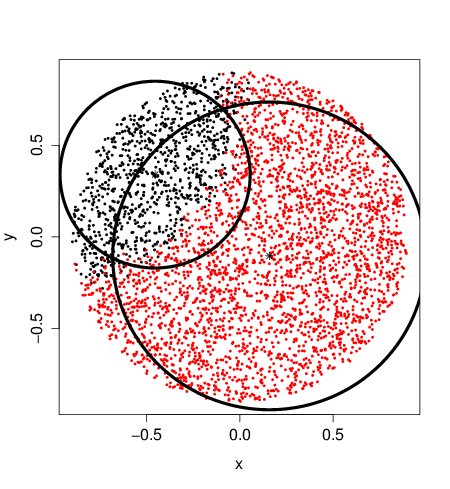}
\includegraphics[width=0.15\textwidth]{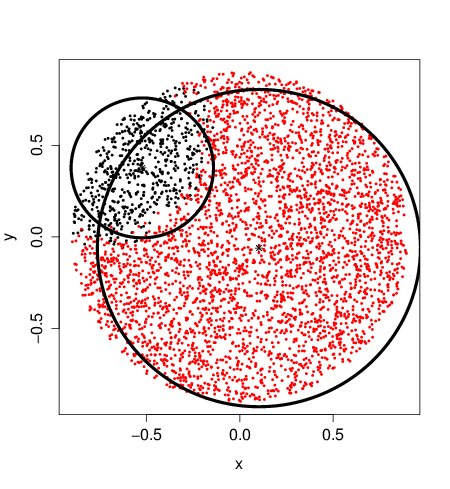}
\includegraphics[width=0.15\textwidth]{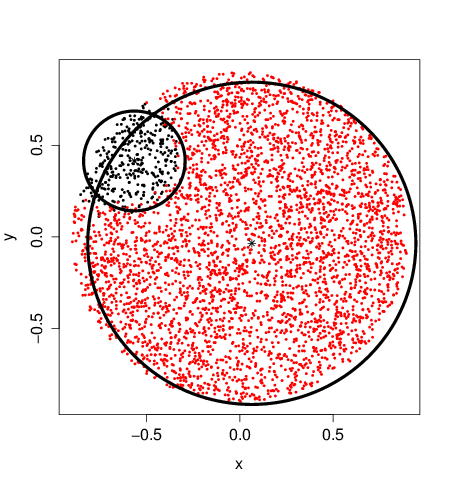}
\includegraphics[width=0.15\textwidth]{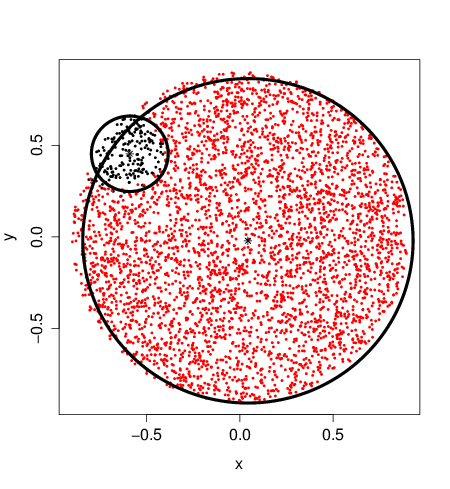}
\includegraphics[width=0.15\textwidth]{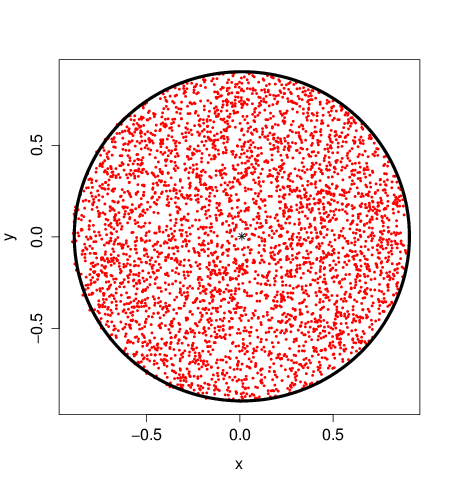}
\caption{The step-by-step view of clusters reduction in the case of a disc--like set.} 
\label{fig:znika} 
\end{figure}

CEC allows an automatic reduction of ``unnecessary'' clusters, since, contrary to the case of classical $k$-means and EM, there is a cost of using each cluster.
To visualize this theory let the results of Gaussian CEC be considered, given in Figure \ref{fig:mouse_5}, where the process started with $k=10$ initial 
randomly chosen clusters which were reduced automatically by the algorithm (used with Spherical CEC).
The step-by-step view of this process can be seen in Figure \ref{fig:znika},
in which the subsequent steps of the Spherical CEC
on data distributed uniformly inside the circle, and divided initially at two almost equal parts are illustrated. 

There are several probabilistic methods which try to estimate the correct number of clusters. For example, \cite{goldberger2004hierarchical} use the generalized distance between Gaussian Mixture Models with different components number by using the Kullback--Leibler divergence,~see \cite{Co-Th, Ku}. A similar approach was presented by \cite{zhang2004competitive} (Competitive Expectation Maximization) which uses the Minimum Message Length criterion provided by \cite{figueiredo2002unsupervised}. 
In practice, MDLP can also be directly used in clustering, see
\cite{wallace1990finding}. However, most of the above mentioned methods typically proceed through all the consecutive clusters and do not reduce
the number of clusters on-line during the clustering process.


For the convenience of the reader, the
contents of the article are hereby briefly summarized.
In the next section a short introduction to the CEC algorithm is provided. Formulas for the cross entropy (which corresponds to the maximal likelihood estimation) of the studied data with respect to the given Gaussian model are also given. The third section concerns parameter fitting and the properties of different Gaussian models. A model with different type of clusters models (``mixed CEC'') is presented as well.
In the last section, the structure of the \proglang{R} Package \pkg{CEC} is presented and  assorted examples are provided.


\section{Theoretical background of CEC}

Let it be recalled that in general EM aims to find 
\begin{equation} \label{eq:cond_p}
p_1,\ldots,p_k \geq 0:\sum_{i=1}^k p_i=1,
\end{equation}
and $f_1,\ldots, f_k \in \F$, where $\F$ is a fixed
(usually Gaussian) family of densities such that the convex combination
\begin{equation} \label{eq:dens}
f:=p_1 f_1 +\ldots p_k f_k
\end{equation}
optimally approximates the scattering of the data under consideration $X=\{x_1,\ldots,x_n\}$. 
The optimization is taken with respect to an MLE based\footnote{Since in clustering the aim is typically to minimize the cost function, the function in \eqref{eq1} is the MLE with a changed sign.} cost function 
\begin{equation} \label{eq1}
\mle(f,X):=-\frac{1}{|X|}\sum_{j=1}^n \ln(f(x_j))=-\frac{1}{|X|}\sum_{j=1}^n \ln\big(p_1 f_1(x_j) +\ldots +p_k f_k(x_j)\big),
\end{equation}
where $|X|$ denotes the cardinality of a set $X$.

The optimization in EM consists of the Expectation and Maximization steps. While the Expectation step is relatively simple, the Maximization
usually (except for the simplest case when the family $F$ 
denotes all Gaussian densities) needs a complicated numerical optimization.

The goal of CEC is similar, i.e. aims at minimizing the
cost function (which is a small modification of that given in \eqref{eq1}
by substituting the sum with a maximum):
\begin{equation} \label{eq2}
\cec(f,X):=-\frac{1}{|X|}\sum_{j=1}^n \ln \big(\max(p_1 f_1(x_j),\ldots,p_k f_k(x_j))\big),
\end{equation}
where all  $p_i$ for $i = 1, \ldots, k$ satisfy the condition \eqref{eq:cond_p}.
It occurs, see \cite{tabor2014cross}, that the above formula implies that, contrary to EM, it is profitable to reduce some clusters (as
each cluster has its cost). Consequently, after minimization for some parameters $i \in \{1,\ldots,k\}$ the 
probabilities $p_i$ will typically equal zero, which implies that the clusters they potentially represent have disappeared. 
Thus $k$, contrary to the case of EM, does not denote the final
number of clusters obtained, but is only an upper bound of the number of clusters of interest (from a series of experiments the authors discovered that typically the good initial guess is to set $k=10$).
Instead of focusing on the density estimation as its
first aim, CEC concerns the clustering, where, similarly to EM, the point $x$ is assigned to the cluster $i$ which maximizes the value $p_if_i(x)$.

However, given the solution to CEC, a good estimation of the initial density is obtained by applying the formula \eqref{eq:dens}, which in practical cases
is very close (with respect to the MLE cost function given by \eqref{eq1}) to the one constructed by EM.

%
%


Let it be remarked that the seemingly small difference in the 
cost function between \eqref{eq1} and \eqref{eq2}
has profound further consequences, which follow
from the fact that the densities in \eqref{eq2} do not ``cooperate'' to build the final approximation of $f$. 

The general idea of cross-entropy clustering relies on finding the splitting of $X \subset \R^N$ into pairwise disjoint sets $X_1,\ldots,X_k$ such that the overall inner information cost of clusters, given in \eqref{eq2}, is minimal.
Consequently, to explain CEC, the cost function to minimize needs to be introduced. To do so, let it be recalled that by the {\em cross-entropy of data set $X$ with respect to density $f$} is given by 
$$
H^{\times}(X\|f) = -\frac{1}{|X|} \sum_{\x \in X} \ln(f(\x)).
$$

Thus, using the information theory approach based on differential entropy (\cite{Co-Th})
instead of the statistical (MLE) point of view, the value of $-\ln f(x)$ in the above sum may be interpreted as the length of code of $x$ with respect to the
coding $f$. In the case of splitting of $X \subset \R^N$
into $X_1, \ldots , X_k$ such that
elements of $X_i$ are ``coded'' by density $f_i$, it can be proven (following \cite{tabor2014cross}) that the mean code--length of a randomly chosen element 
$x \in X$ equals
\begin{equation}\label{en:cec}
\cec(X_1,f_1; \ldots; X_k,f_k ):= \sum_{i=1}^{k} p_i  \cdot \left( -\ln(p_i) + H^{\times}(X_i\|f_i) \right),
\text{ where }p_i = \tfrac{|X_i|}{|X|}.
\end{equation}

Roughly speaking, the first component $-\ln(p_i) $ in the brackets on the RHS is the number of Nats necessary to identify which algorithm is used for coding the element $x \in X_i$ and the second one, $H^{\times}(x\|f_i)$, is the mean code-length of coding $X_i$ by the density $f_i$. Thus, the use of each cluster 
is in a natural way penalized by the function $-\ln(p_i)$
(the cost of identifying the cluster), which consequently causes
the reduction of those clusters which do not add
to the total quality of clustering.

To efficiently use mixture models in CEC, only the 
optimal value of the coding of $X$ needs to be computed:
$$
H^{\times}(X\|\F) := \inf_{f \in \F} H^{\times}(X\|f)
$$
with respect to the density family $\F$. 

\medskip

\noindent{\textbf{Optimization condition.}} \newline
Summarizing, given the density families $\F_1,\ldots,\F_n$, the goal of the CEC algorithm is to divide the data-set $X$ into $k$ (possibly empty) clusters $X_1,\ldots,X_k$ such that the value of the function 
\begin{equation}\label{en:cecfam}
\cec(X_1,\F_1;\ldots;X_k,\F_k):=\sum_{i=1}^k p_i  \cdot \left( -\ln(p_i) + H^{\times}(X_i\|\F_i) \right),
\text{ where }p_i = \frac{|X_i|}{|X|}
\end{equation}
is minimal.

In practice the exact formula
for the value of cross-entropy $H^{\times}(X,\F)$
for the most common subfamilies $\F$ of all Gaussian densities can easily be derived, see \cite{tabor2014cross,Ta-Mi}  (in the
following pages, the formula for the
six most commonly encountered Gaussian subfamilies are given). Since each Gaussian is uniquely identified by its mean and covariance, the denotation for the estimators of mean
and covariance of the random variable is needed, the realization of which
is given by the data set $X$. As usual, by an estimator of the mean and covariance we take
$$
\m_X := \frac{1}{|X|}\sum \limits_{\x \in X} \x,
$$ 
$$ 
\Sigma_X := \frac{1}{|X|} \sum \limits_{\x \in X} (\x-\m_X)(\x-\m_X)^\top.
$$

The ground is now set to present the exact formula for the cross-entropy of Gaussian subfamilies implemented in the \pkg{CEC} Package:

\hspace{1cm}

1. $\quad \G_{\Sigma}$ -- Gaussian densities with covariance $\Sigma$. The 
clustering will have the tendency to divide the data into clusters resembling balls with respect to the Mahalanobis distance $\|\cdot\|_{\Sigma}$.
\begin{tabular}{ | m{4cm}  p{10cm} |}
\hline
  \qquad\includegraphics[trim=0 0 0 -5,width= 0.10\textwidth]{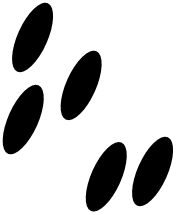} 
  & $\Sigma_{\G_{\Sigma}}(X)  =\Sigma $  \newline 
  	$H^{\times}(\y\|\G_{\Sigma})=\frac{N}{2} \ln(2\pi)+\frac{1}{2}\tr(\Sigma^{-1}\Sigma_{X})+\frac{1}{2}\ln \det(\Sigma) $ \\
\hline
\end{tabular}

\hspace{1cm}

2. $\quad \G_{r\I}$ -- subfamily of $\G_{\Sigma}$,
for $\Sigma=r\I$ and $r>0$ is fixed, which 
consists of the spherical (radial Gaussian) with  
covariance matrix $r \I$ (the clustering will have tendency to divide the data into balls with fixed radius proportional to $\sqrt{r}$).  

\begin{tabular}{ | m{4cm}  p{10cm} |}
\hline
  \qquad\includegraphics[trim=0 0 0 -5,width= 0.10\textwidth]{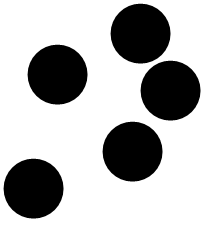} 
  & $\Sigma_{\G_{r\I}}(X)  =r \I   $  \newline 
  $H^{\times}(X\|\G_{r\I})=\frac{N}{2}\ln(2\pi )+\frac{N}{2}\ln (r) + \frac{1}{2r}\tr(\Sigma_X)$
  	\\
\hline
\end{tabular}    

\hspace{1cm}

3. $\quad \G_{(\cdot I)}$ -- spherical (radial) Gaussian densities meaning those Gaussians for which the covariance is
proportional to identity. The clustering will try to divide the data into balls of arbitrary sizes.

\begin{tabular}{ | m{4cm}  p{10cm} |}
\hline
  \qquad\includegraphics[trim=0 0 0 -5,width= 0.11\textwidth]{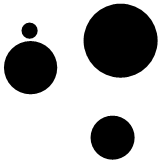} 
  & $\Sigma_{\G_{(\cdot I)}}(X)  =\frac{\tr(\Sigma_{X})}{N} \I  $  \newline 
  $H^{\times}(X\|\G_{(\cdot I)})=\frac{N}{2}\ln(2\pi e/N)+\frac{N}{2}\ln (\tr \Sigma_X) $
  	\\
\hline
\end{tabular}    

\hspace{1cm}

4. $\quad \G_{\mathrm{diag}}$ -- Gaussians with diagonal covariance. The  clustering will try to divide the data into ellipsoids with radii parallel to coordinate axes.

\begin{tabular}{ | m{4cm}  p{10cm} |}
\hline
  \qquad\includegraphics[trim=0 0 0 -5,width= 0.15\textwidth]{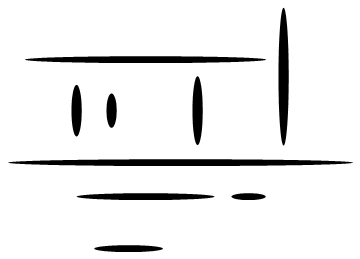} 
  & $\Sigma_{\G_{\mathrm{diag}}}(X)  =\mathrm{diag}(\Sigma_{X})   $ \newline 
  $H^{\times}(X\|\G_{\mathrm{diag}})=\frac{N}{2}\ln(2\pi e)+\frac{1}{2}\ln(\det(\mathrm{diag}(\Sigma_X)))  $
  	\\
\hline
\end{tabular}     

\hspace{1cm}  

5. $\quad \G_{\lambda_1,\ldots,\lambda_N}$ -- Gaussian densities with the covariance matrix having eigenvalues $\lambda_1,\ldots,\lambda_N$ such that $\lambda_1\leq\ldots\leq\lambda_N$. The clustering will try to divide the data into ellipsoids with fixed shape rotated by an arbitrary angle.

\begin{tabular}{ | m{4cm}  p{10cm} |}
\hline
  \qquad\includegraphics[trim=0 0 0 -5,width= 0.15\textwidth]{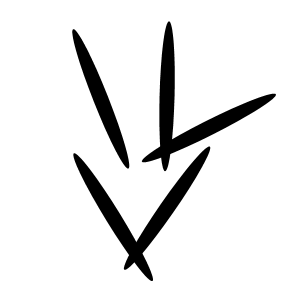} 
  & $ \Sigma_{\G}(X)=\Sigma_{{\lambda_1,\ldots,\lambda_N}} $ \newline 
  $H^{\times}(X\|\G_{\lambda_1,\cdots,\lambda_N})=\frac{N}{2} \ln(2\pi)+\frac{1}{2}
\sum_{i=1}^N \frac{\lambda_i^{X}}{\lambda_i}+\frac{1}{2}\ln \bigg(\prod_{i=1}^N\lambda_i\bigg)$
  	\\
\hline
\end{tabular}      
         
\hspace{1cm}  
  
6. $\quad \G$ -- all Gaussian densities. In this case the dataset is divided into ellipsoid-like clusters without any preferences concerning the size or the shape of the ellipsoid.

\begin{tabular}{ | m{4cm}  p{10cm} |}
\hline
  \qquad\includegraphics[trim=0 0 0 -5,width= 0.15\textwidth]{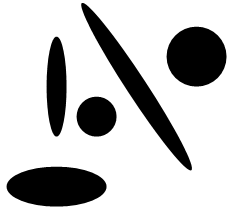} 
  & $ \Sigma_{\G}(X)=\Sigma_{X} $ \newline 
  $H^{\times}(X\|\G)=\frac{N}{2}\ln(2\pi e)+\frac{1}{2}\ln \det(\Sigma_{X}) $
  	\\
\hline
\end{tabular}


\vspace{1cm} 
\textbf{The algorithm behind CEC} \newline

In this subsection the basic information about the implementation of CEC in \proglang{R} is presented. As usual in clustering the process is started with the initialization of the clusters, which can either be done by choosing centers from the dataset randomly  and assigning points to the nearest center, or by the $k$-means++  approach proposed by \cite{kmeans++}.

Since CEC, from both the implementation and the theoretical point of view, is 
a generalization of $k$-means, in the search for the minimum of the cost function the typical approaches used in $k$-means can be used -- the Lloyd's and the Hartigan's methods. From the practical point of view, Hartigan's approach finds smaller minima and reduces the unnecessary clusters in a better way, but at the cost of
recomputing the covariance at each passing through every data point. Consequently, for the low-dimensional data the authors suggest using the Hartigan's, while for high dimensional the
Lloyd's method. 

The idea of the Hartigan method is to proceed over all elements of $X$, and switch the membership to those clusters which would maximally decrease the cost function. Since in the discussed approach the clusters are removed, the classical Hartigans approach is slightly modified.

\medskip
\textbf{Hartigan's procedure} \newline
Consider $k$ sub density families\footnote{Only the density family for which formulas are presented at the end of the previous section is used.} $(\F_i)_{i=1}^k$.
To explain the Hartigan approach more precisely the notion of {\em cluster membership function} is needed
$$
\cl \colon X \to \{0,\ldots,k\},
$$
where element $x \in X$ belongs to the 
$\cl(x)$-th cluster: $X_{\cl(x)}$ ($0$ is reserved
as a special symbol which denotes the fact that 
$x$ is unassigned).

Such a cluster membership function $\cl \colon X \to \{1,\ldots,k\}$ is desired (thus all elements of $X$ are assigned) that the value of 
$$
\cec(X_1,\F_1;\ldots;X_k,\F_k),
\text{ where }X_l:=\{x \in X: \cl(x)=l\}=\cl^{-1}(l),
$$ 
 is minimal. 

The basic idea of Hartigan is relatively simple -- the process goes over all elements of $X$ and the following steps are applied:
\begin{itemize}
\item if the chosen $x \in X$ is unassigned, assign it to the arbitrary nonempty cluster;
\item reassign $x$ to those clusters for which the decrease in cross-entropy is maximal;
\item check if no cluster needs to be removed\footnote{The given cluster is usually removed if it falls below some percentage level of all data, which was usually fixed at $5\%$.}, if this is the case remove its all elements;
\end{itemize}
until no cluster membership has been changed
during the whole iteration over set $X$.

Observe that when dealing with Gaussian families discussed in the previous section to compute $H^{\times}(X_i\|\F_i)$, the cardinality of $X_i$ and its covariance need to be known. This implies that in practice the whole cluster $X_i$ does not need to be remembered -- it is sufficient to know its covariance and cardinality.

It occurs that in practice, after adding or deleting point $x$ to the cluster, the covariance 
and cardinality of $X_i$ can be updated on-line. Therefore, only the value of the mean and the covariance matrix of $X_i$ needs to be remembered.

This is discussed in the following observation ($Y_1$ plays
the role of cluster $X_i$, and $Y_2$ denotes by default
the point $x$ which we either add or remove from cluster $X_i$). Consider sets $Y_1,Y_2 \subset \R^N$:

a) The case\footnote{This corresponds to the case when the point $x$ is added to the cluster $X_i$ .} when $Y_1 \cap Y_2=\emptyset$ is first discussed. Then 
$$
\begin{array}{l}
\m_{Y_1 \cup Y_2} =  p_{1}\m_{Y_1}+p_{2}\m_{Y_2}, \\[1ex]
\Sigma_{Y_1 \cup Y_2} =  
p_{1}\Sigma_{Y_1}+p_2\Sigma_{Y_2}
+p_{1}p_{2}(\m_{Y_1}-\m_{Y_2})(\m_{Y_1}-\m_{Y_2})^T,
\end{array}
$$
where $p_1=\frac{|Y_1|}{|Y_1| + |Y_2|}$ and $p_2=\frac{|Y_2|}{|Y_1|+|Y_2|}$.

b) Assume\footnote{This corresponds to the case when the point $x$ is removed from the cluster $X_i$.} that $Y_1 \subsetneq Y_2$.  Then
$$
\begin{array}{l}
\m_{Y_1 \setminus Y_2} =  q_1 \m_{Y_1}-q_2\m_{Y_2}, \\[1ex]
\Sigma_{Y_1 \setminus Y_2}= 
q_1\Sigma_{Y_1}-q_2\Sigma_{Y_2}
-q_1 q_2 (\m_{Y_1}-\m_{Y_2})(\m_{Y_1}-\m_{Y_2})^T,
\end{array}
$$
where $q_1:=\frac{|Y_1|}{|Y_1|-|Y_2|}$ and  $q_2:=\frac{|Y_2|}{|Y_1|-|Y_2|}$.

From the above equations formula in the case of adding one point and removing some of them can easily be obtained. 

\section{The CEC package}

\begin{figure}[!t] \centering
	\includegraphics[width=0.7\textwidth]{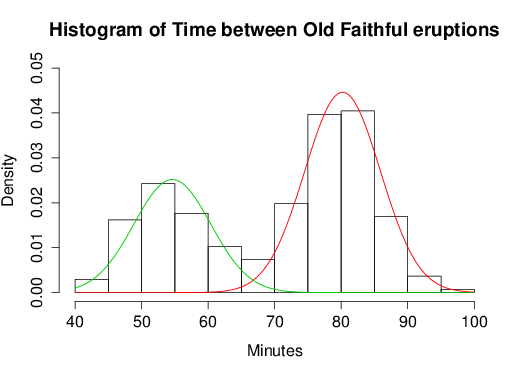}
	\caption{The Old Faithful waiting data fitted with a CEC model.}
	\label{fig:R1Gauss} 
\end{figure}

In this section the implementation of the CEC algorithm in the \proglang{R} Package \pkg{CEC} is presented.
Consider first the Old Faithful data in $\R$, see \cite{azzalini1990look}. Observe that the data distribution resembles a mixture of two  Gaussians, see Figure \ref{fig:R1Gauss}. 

In the basic use of this package the input dataset \code{data} and the initial number \code{centers} of clusters: \code{cec(x = ..., centers = ...)} have to be specified. Below, a simple session with \proglang{R} is presented, where the component
(waiting) of the Old Faithful dataset is split into two clusters.

\begin{CodeChunk}
\begin{CodeInput}
R> library("CEC")
R> attach(faithful)
R> cec <- cec(matrix(faithful$waiting), 2)
R> print(cec)
\end{CodeInput}
\begin{CodeOutput}
CEC clustering result: 

Clustering vector: 
  [1] 1 2 1 2 1 2 1 1 2 1 2 1 1 2 1 2 2 1 2 1 2 2 1 1 1 1 2 1 1 1 1 1 2 1
 [35] 1 2 2 1 2 1 1 2 1 2 1 1 2 2 1 2 1 1 2 1 2 1 1 2 1 1 2 1 2 1 2 1 1 1
 [69] 2 1 1 2 1 1 2 1 2 1 1 1 1 1 1 2 1 1 1 1 2 1 2 1 2 1 2 1 1 1 2 1 2 1
[103] 2 1 1 2 1 2 1 1 1 2 1 1 2 1 2 1 2 1 2 1 1 2 1 1 2 1 2 1 2 1 2 1 2 1
[137] 2 1 2 1 1 2 1 1 1 2 1 2 1 2 1 1 2 1 1 1 1 1 2 1 2 1 2 1 2 1 2 1 2 1
[171] 2 2 1 1 1 1 1 2 1 1 2 1 1 1 2 1 1 2 1 2 1 2 1 1 1 1 1 1 2 1 2 1 1 2
[205] 1 2 1 1 2 1 1 1 2 1 2 1 2 1 2 1 2 1 2 1 1 1 1 1 1 1 1 2 1 2 1 2 2 1
[239] 1 2 1 2 1 2 1 1 2 1 1 1 2 1 1 1 1 1 1 1 2 1 1 1 2 1 2 2 1 1 2 1 2 1

Probability vector:
[1] 0.6360294 0.3639706

Means of clusters:
         [,1]
[1,] 80.20809
[2,] 54.62626

Cost function at each iteration:
[1] 3.820302 3.817422 3.817422

Number of clusters at each iteration:
[1] 2 2 2

Number of iterations:
[1] 2

Computation time:
[1] 0

Available components:
 [1] "data"                "cluster"             "probabilities"      
 [4] "centers"             "cost.function"       "nclusters"          
 [7] "final.cost.function" "final.nclusters"     "iterations"         
[10] "covariances"         "covariances.model"   "time" \end{CodeOutput}
\end{CodeChunk}

As its main outcome \pkg{CEC} returns data cluster membership \code{cec$cluster}, which corresponds to the function $\cl \colon X \to \{1, \ldots,k \}$ from the previous section. The following parameters of 
clusters $X_1,\ldots,X_k$ are obtained as well:
\begin{itemize}
\item $p_i=\tfrac{|X_i|}{|X|}$ (probabilities of clusters);
\item $\m_i$: means of clusters;
\item $\Sigma_i$: covariances of clusters.
\end{itemize}
The above are necessary to obtain the calculated subdensity
$$
\max\left(p_1 \cdot N_{(\mu_1,\Sigma_1)},\ldots, p_k \cdot N_{(\mu_k,\Sigma_k)}\right),
$$
which can be used to compute the cost function (given
by the cross-entropy) or the identification of the cluster
membership of new points (a point $\x$ belongs to this
cluster where the value $N_{(\mu_i,\Sigma_i)}(x)$
is maximized). Moreover, the above can be used to 
compute the density estimation 
$$
p_1 \cdot N_{(\mu_1,\Sigma_1)}+\ldots+p_k \cdot N_{(\mu_k,\Sigma_k)}.
$$

The parameters of the CEC model are stored as: 
\begin{enumerate}
\item a list of means (\code{cec$centers}, namely $\mu_i$ for $i=1,\ldots,k$), 
\item a list of covariances (\code{cec$covariances.model}, namely $\Sigma_i$ for $i=1,\ldots,k$), 
\item a list of probabilities (\code{cec$probability}, namely $p_i$ for $i=1,\ldots,k$). 
\end{enumerate}
Some additional information concerning the number of iterations, cost (energy) function and the number of clusters during the following iterations is also obtained.  

Below, a session of \proglang{R} is presented which shows how to use the above parameters for plotting the data and the Gaussian models corresponding to the clusters.

\begin{CodeChunk}
\begin{CodeInput}
R> hist(faithful$waiting, prob = TRUE, main = "Histogram of Time between Old
+    Faithful eruptions", xlab = "Minutes", ylim = c(0, 0.05));
R> for(i in c(1:2)){
R> curve(cec$probability[i] * dnorm(x, mean = cec$centers[i], 
+    sd = sqrt(cec$covariances[[i]][1])), add = T, col = i + 1)  
R> }
\end{CodeInput}
\end{CodeChunk}

As it was said, the discussed method, analogously to $k$-means, depends on the initial clusters memberships. Therefore, the initialization should be started a few times, which can be obtained with the use of parameter \code{nstart} (e.g., \code{cec <- cec(x = ...,  centers = ..., nstart = ...)}). 
The initial cluster membership function can be chosen by the use of \code{centers.init} either randomly, \code{"random"}, or with the method given by the $k$-means++ algorithm (\cite{kmeans++}), \code{"kmeans++"}. 

Two more parameters are important in the initialization. The first \code{iter.max = 100} equals the maximum number of iterations in one CEC start and the second  \code{card.min = "5\%"} is the percentage of the minimal size of each cluster. The second parameter specifies the minimal possible number points in each cluster (clusters which contains less points are removed). Since each cluster is described by a covariance matrix, the number of elements in the cluster must be larger than the dimension of the data. 

One of the most important properties of the CEC algorithm is that it can be applied for various Gaussian models. Therefore, the \pkg{CEC} package includes the implementation of six Gaussian models, which can be specified by the parameter \code{type}.
All the models implemented in the \pkg{CEC} package are discussed below.

\begin{figure}[!t] \centering
	\subfigure[Randomly generated four Gaussians dataset.]{\label{fig:vor_1_a}\includegraphics[width=0.32\textwidth]{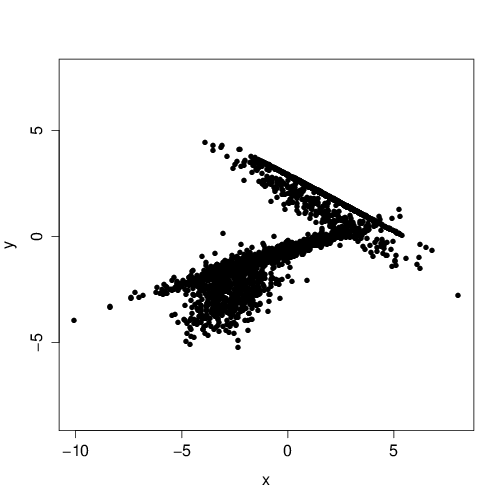}}
	\subfigure[Effect of CEC with an initial number of $k=10$ clusters.]{\label{fig:vor_2_a}\includegraphics[width=0.32\textwidth]{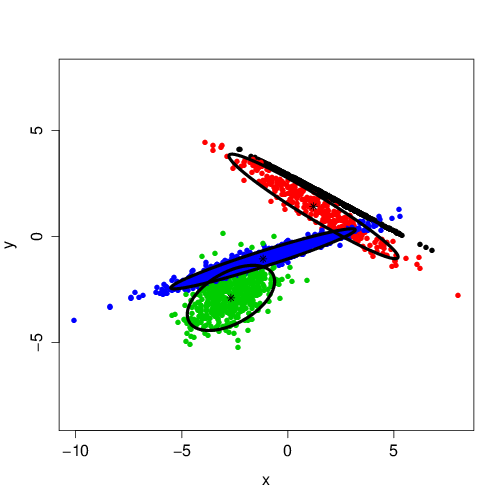}}
	\subfigure[Decrease of the cost function.]{\label{fig:vor_2_b}\includegraphics[width=0.32\textwidth]{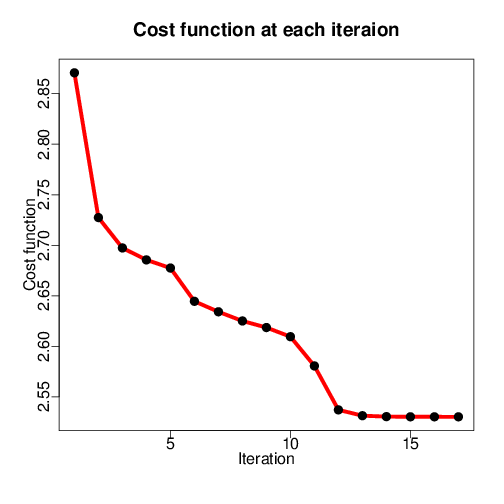}}
	\caption{Clustering with respect to the general Gaussian model. Figure \protect\subref{fig:vor_2_b} presents the decrease of the cost function in time.}
	\label{fig:fourGauss} 
\end{figure}

\medskip
\textbf{$\G$ -- General Gaussian distributions} \newline

The family containing all Gaussian distributions $\G$ is considered first. 
The results of the general Gaussian CEC algorithm give similar results to those obtained  by the Gaussian Mixture Models. 
However, the authors' method does not use the EM (Expectation Maximization) approach for minimization but a simple iteration process (Hartigan method). Consequently,  larger datasets can be processed in shorter time.

The clustering will have the tendency to divide
the data into clusters in the shape of ellipses
(ellipsoids in higher dimensions). 
 
\begin{CodeChunk}
\begin{CodeInput}
R> library("CEC")
R> data("fourGaussians")
R> cec <- cec(fourGaussians, centers = 10, type = "all", nstart = 20)
R> plot(cec, xlim = c(0, 1), ylim = c(0, 1), asp = 1)
R> cec.plot.cost.function(cec)
\end{CodeInput}
\end{CodeChunk}
  
It can be used for exploring the data structure in the case when no information about the relations in the dataset is available. After the analysis of the outcome, the decision can be made to use more specific types of Gaussian families.

The result of CEC algorithms with various types of Gaussian models on T type sets are presented in Fig. \ref{fig:cecmod}. The Figure was generated by the following codes in \proglang{R}:

\begin{figure}[!h] \centering
		\subfigure[The T-dataset.]{\label{fig:cecmod_1}\includegraphics[width=0.35\textwidth]{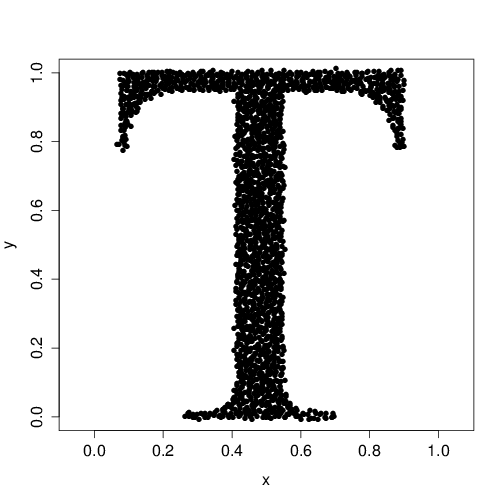}} 
		\subfigure[Effect of the spherical CEC.]{\label{fig:cecmod_2}\includegraphics[width=0.35\textwidth]{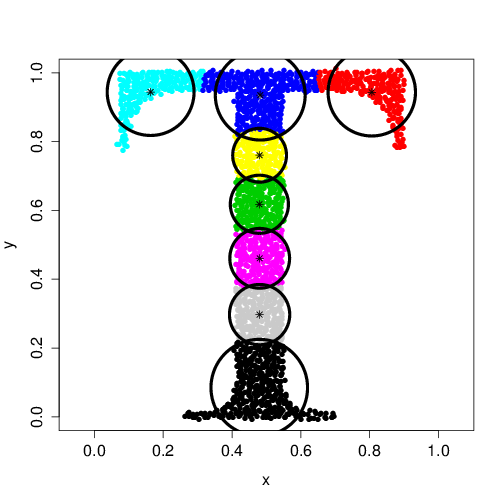}}
		\subfigure[Effect of the spherical CEC with fixed radius.]{\label{fig:cecmod_3}\includegraphics[width=0.35\textwidth]{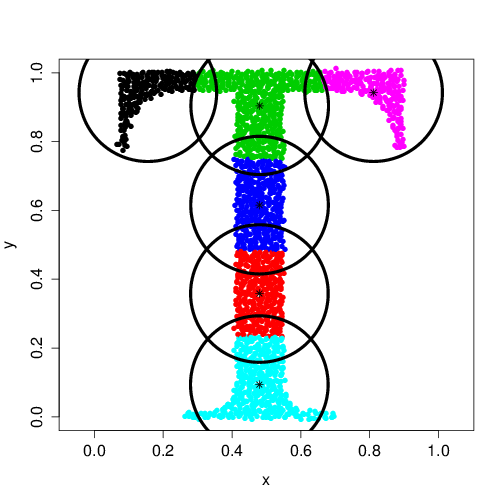}}		
		\subfigure[Effect of the diagonal CEC.]{\label{fig:cecmod_4}\includegraphics[width=0.35\textwidth]{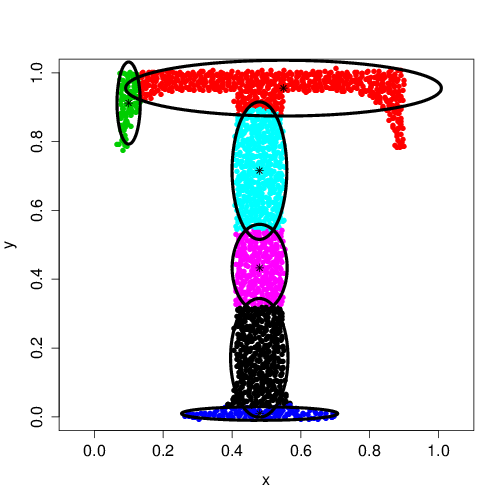}}
		\subfigure[Effect of the fixed covariance CEC.]{\label{fig:cecmod_5}\includegraphics[width=0.35\textwidth]{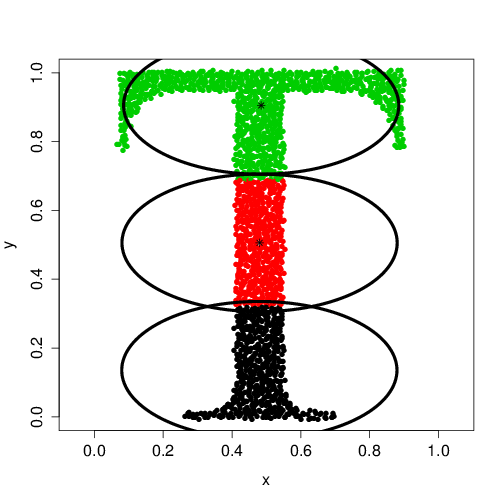}}
				\subfigure[Effect of the fixed eigenvalue CEC.]{\label{fig:cecmod_6}\includegraphics[width=0.35\textwidth]{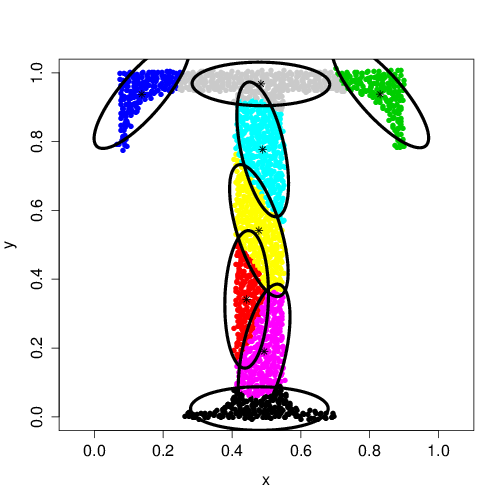}}
		\caption{The CEC algorithm in the case of clustering a T-type set according to the various types of the CEC model.}
	\label{fig:cecmod} 
\end{figure}

\begin{CodeChunk}
\begin{CodeInput}
R> library("CEC")
R> data("Tset")
\end{CodeInput}
\end{CodeChunk}
\begin{itemize}
\item spherical CEC:
\begin{CodeChunk}
\begin{CodeInput}
R> cec <- cec(x = Tset, centers = 10, type = "spherical")
R> plot(cec, xlim = c(0, 1), ylim = c(0, 1), asp = 1)
\end{CodeInput}
\end{CodeChunk}
\item spherical CEC with fixed radius:
\begin{CodeChunk}
\begin{CodeInput}
R> cec <- cec(x = Tset, centers = 10, type = "fixedr", param = 0.01)
R> plot(cec, xlim = c(0, 1), ylim = c(0, 1), asp = 1)
\end{CodeInput}
\end{CodeChunk}
\item diagonal CEC:
\begin{CodeChunk}
\begin{CodeInput}
R> cec <- cec(x = Tset, centers = 10, type = "diagonal")
R> plot(cec, xlim = c(0, 1), ylim = c(0, 1), asp = 1)
\end{CodeInput}
\end{CodeChunk}
\item fixed covariance CEC:
\begin{CodeChunk}
\begin{CodeInput}
R> cec <- cec(x = Tset, centers = 10, type = "covariance", 
+    param = matrix(c(0.04, 0, 0, 0.01), 2))
R> plot(cec, xlim = c(0, 1), ylim = c(0, 1), asp = 1)
\end{CodeInput}
\end{CodeChunk}
\item fixed eigenvalue CEC:
\begin{CodeChunk}
\begin{CodeInput}
R> cec <- cec(x = Tset, centers = 10, type = "eigenvalues", 
+    param=c(0.01, 0.001))
R> plot(cec, xlim = c(0, 1), ylim = c(0, 1), asp = 1)
\end{CodeInput}
\end{CodeChunk}
\end{itemize}

\medskip
\textbf{$\G_{(\cdot I)}$ -- Spherical Gaussians} \newline

The second family discussed contains spherical Gaussian distributions $\G_{(\cdot I)}$ which 
can be accessed by
\code{cec(x = ..., centers = ..., type = "spherical")}.
The original distribution will be estimated by spherical (radial) densities, which will result with splitting the data into circle-like clusters of arbitrary
sizes (balls in higher dimensions).

In Fig. \ref{fig:cecmod_2} the result of the spherical algorithm with circles fitted to the obtained clusters is presented. 
This family can be used for the recognition of circular--shape objects, see \cite{Sp-Ta}.

\medskip
\textbf{$\G_{r\I}$ -- Spherical Gaussians with a fixed radius} \newline

The next model implemented in the \pkg{CEC} package is a spherical model with a fixed covariance: 
\code{cec(x = ..., centers = ..., type = "fixedr", param = ...)}. 
Similarly to the general spherical model, the dataset will be divided into clusters resembling full circles, but with the radius determined by \code{param}.

In Fig. \ref{fig:cecmod_3} the result of the spherical fixed radius algorithm with ellipses fitted to the obtained clusters is presented.

\medskip
\textbf{$\G_{\mathrm{diag}}$ -- Diagonal Gaussian} \newline

The fourth model is based on diagonal Gaussian densities (e.g \code{cec(x = ..., centers = ..., type = "diagonal")}). In this case, the data will be described by ellipses for which the main semi-major axes are parallel to the axes of the coordinate system. In Fig. \ref{fig:cecmod_4} the result of the spherical fixed radius algorithm with ellipses fitted to the obtained clusters is presented. 


\medskip
\textbf{$\G_{\Sigma}$ -- Gaussian with fixed covariance} \newline

The next model contains Gaussians with an arbitrary fixed covariance matrix  e.g \newline
 \code{cec(x = ..., centers = ..., type = "covariances", param = ...)}.
In this example
$
\begin{bmatrix}
0.04 & 0 \\
0 & 0.01
\end{bmatrix}
$
is used, which means that the data is covered by fixed ellipses.
In Fig. \ref{fig:cecmod_5} the result of the fixed covariance CEC is presented.

\medskip
\textbf{$\G_{\lambda_1,\ldots,\lambda_N}$ -- Gaussian densities with fixed eigenvalues $\lambda_1,\ldots,\lambda_N$} \newline

The last model is based on Gaussians with arbitrary fixed eigenvalues  (e.g \code{cec(x = ..., centers = ..., type = "eigenvalues", param = ...)}). In this example $\lambda_1=0.01$, $\lambda_2=0.001$ are used, which means that the set is covered by ellipses with fixed semi axes (which correspond to the eigenvalues). 
In Fig. \ref{fig:cecmod_6} the result of the fixed eigenvalues CEC is presented.

\begin{figure}[!h] \centering
		%
		\subfigure[Iris dataset.]{\label{fig:cecmod_6}\includegraphics[width=0.58\textwidth]{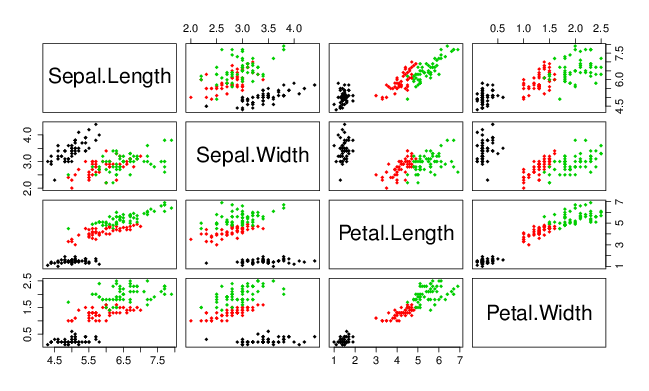}}
		\subfigure[Wine dataset.]{\label{fig:cecmod_6}\includegraphics[width=0.38\textwidth]{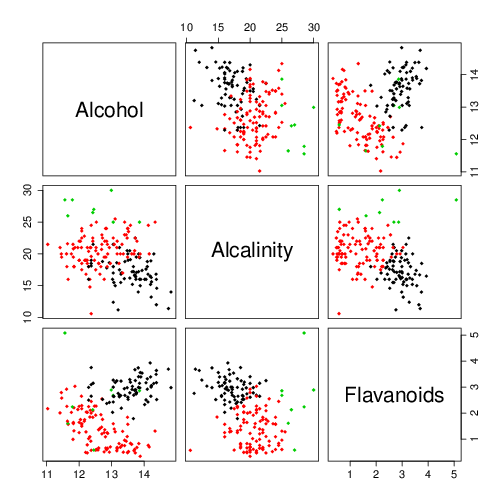}}
		%
		%
		%
		\caption{Clustering of iris and wine datasets according to the general CEC model.}
	\label{fig:cecmod2} 
\end{figure}

At the end of this section we present how our method works on data from UCI repository. In the first example we consider iris dataset, which consists of 50
samples from each of three classes of iris flowers \cite{fisher1936use}. One class is linearly separable from the other two, while the latter are not linearly separable from each other, see Fig. \ref{fig:cecmod2}. Next we consider three coordinates of wine data set, analogically to experiments from introduction to the \proglang{R} package \pkg{pdfCluster} \cite{Azzalini2013}. The wine data set was introduced by \cite{forina1986multivariate}. It originally included the results of 27 chemical measurements on 178 wines grown in the same region in Italy but derived from three different cultivars: Barolo, Grignolino and Barbera, see Fig. \ref{fig:cecmod2}.

\section[Gaussian]{A mix of the Gaussian models}\label{cha:mix}

One of the most powerful properties of the CEC algorithm is the possibility of mixing models. More precisely, the mixed models can be specified by giving a list of cluster types 

\code{cec(x = ..., centers = ..., type = c("all", ...), param = ...)}. 

\begin{figure}[!h] \centering
		\subfigure[Dataset containing two types of patterns (circular and elliptical).]{\label{fig:cecmod_a}\includegraphics[width=0.4\textwidth]{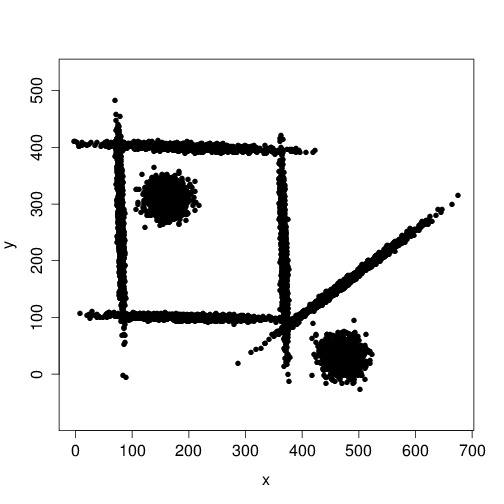}}
		\subfigure[Result of CEC with two Gaussians with a fixed radius and four with fixed eigenvalues of covariance.]
{\label{fig:cecmod_a}\includegraphics[width=0.4\textwidth]{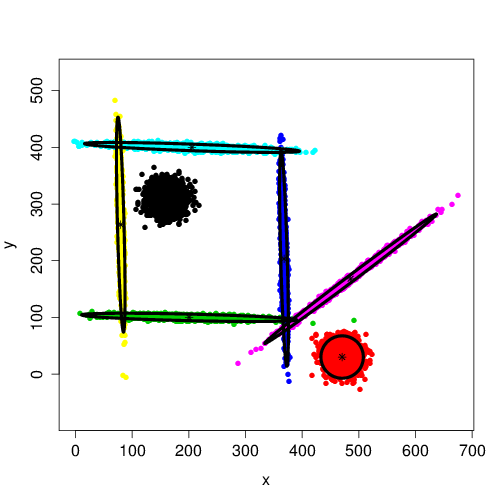}}
		\caption{The CEC algorithm in the case of clustering according to a mixed model.}
	\label{fig:cecmixed} 
\end{figure}

  Fig. \ref{fig:cecmixed} presents the CEC clustering according to two clusters described by spherical Gaussians with a fixed radius ($\G_{r\I}$) $r=350$ and five clusters of type $\G_{\lambda_1,\ldots,\lambda_N}$ with fixed eigenvalues 
\code{c(9000, 8)}. This kind of configurations can be used in many cases, especially if a wide knowledge of the structure of the investigated set is possessed. Various patterns of the image can be distinguished, for example multiple types of objects can be detected simultaneously, e.g., the search for matches (Gaussian with specified covariance matrix) and coins (spherical Gaussian with fixed radius) is possible at the same time -- compare with \cite{Ta-Mi}.

Figure \ref{fig:cecmixed} was generated by the following code in \proglang{R}:

\begin{CodeChunk}
\begin{CodeInput}
R> library("CEC")
R> data("mixShapes")
R> cec <- cec(mixShapes, 7, type = c("fixedr", "fixedr", "eigen", "eigen", 
+    "eigen", "eigen", "eigen"), param = list(350, 350, c(9000, 8), 
+    c(9000, 8), c(9000, 8), c(9000, 8), c(9000, 8)), nstart = 100)
R> plot(cec, asp = 1)
\end{CodeInput}
\end{CodeChunk}

\section{Concluding remarks}

The \proglang{R}  \pkg{CEC} package proposed in this work uses cross-entropy clustering described by \cite{tabor2014cross}.  The presented method is an interesting alternative to the classical clustering methods like $k$-means, EM, GMM and their generalizations. Since CEC does not use the EM method, new models can be added without the need for using complicated optimization. Another important property of CEC is the automatic reduction of the clusters which have a negative information cost.

The main advantage of the method lies in the fact that it can be easily adapted to different Gaussian models. 
Thus, the package enables to specify which kind of Gaussian subfamilies will be used in clustering. In particular, it is possible to use: spherical Gaussians, spherical Gaussians with the fixed radius, diagonal Gaussians, Gaussians with the fixed covariance or Gaussians with fixed eigenvalues. Moreover, it is possible to use a combination of the above mentioned types of Gaussian subfamilies. 
The package also proposes the tools to visualize the obtained results.

\section{Acknowledgments}

The work of P. Spurek was supported by the National Centre of Science (Poland)
[grant no. 2013/09/N/ST6/01178]. The work of J. Tabort was supported by the National Centre of Science (Poland) [grant no. 2014/13/B/ST6/01792].



\begin{thebibliography}{49}
\newcommand{\enquote}[1]{``#1''}
\providecommand{\natexlab}[1]{#1}
\providecommand{\url}[1]{\texttt{#1}}
\providecommand{\urlprefix}{URL }
\expandafter\ifx\csname urlstyle\endcsname\relax
  \providecommand{\doi}[1]{doi:\discretionary{}{}{}#1}\else
  \providecommand{\doi}{doi:\discretionary{}{}{}\begingroup
  \urlstyle{rm}\Url}\fi
\providecommand{\eprint}[2][]{\url{#2}}

\bibitem[{Arthur and Vassilvitskii(2007)}]{kmeans++}
Arthur D, Vassilvitskii S (2007).
\newblock \enquote{{k-means++: The advantages of careful seeding}.}
\newblock In \emph{Proceedings of the eighteenth annual ACM-SIAM symposium on
  Discrete algorithms}, pp. 1027--1035. Society for Industrial and Applied
  Mathematics.

\bibitem[{Azzalini and Bowman(1990)}]{azzalini1990look}
Azzalini A, Bowman A (1990).
\newblock \enquote{A look at some data on the Old Faithful geyser.}
\newblock \emph{Applied Statistics}, pp. 357--365.

\bibitem[{Azzalini and Menardi(2014)}]{Azzalini2013}
Azzalini A, Menardi G (2014).
\newblock \enquote{Clustering via Nonparametric Density Estimation: The
  \proglang{R} Package \pkg{pdfCluster}.}
\newblock \emph{Journal of Statistical Software}, \textbf{57}(11).
\newblock ISSN 1548-7660.
\newblock \urlprefix\url{http://www.jstatsoft.org/v57/i11}.

\bibitem[{Benaglia \emph{et~al.}(2009)Benaglia, Chauveau, Hunter, and
  Young}]{Benaglia2009}
Benaglia T, Chauveau D, Hunter DR, Young DS (2009).
\newblock \enquote{\pkg{mixtools}: An \proglang{R} Package for Analyzing
  Mixture Models.}
\newblock \emph{Journal of Statistical Software}, \textbf{32}(6).
\newblock ISSN 1548-7660.
\newblock \urlprefix\url{http://www.jstatsoft.org/v32/i06}.

\bibitem[{Berg{\'e} \emph{et~al.}(2012)Berg{\'e}, Bouveyron, and
  Girard}]{Berge2011}
Berg{\'e} L, Bouveyron C, Girard S (2012).
\newblock \enquote{\pkg{HDclassif}: An \proglang{R} Package for Model-Based
  Clustering and Discriminant Analysis of High-Dimensional Data.}
\newblock \emph{Journal of Statistical Software}, \textbf{46}(6), 1--29.
\newblock ISSN 1548-7660.
\newblock \urlprefix\url{http://www.jstatsoft.org/v46/i06}.

\bibitem[{Bock(2007)}]{bock2007}
Bock H (2007).
\newblock \enquote{{Clustering methods: a history of K-Means algorithms}.}
\newblock \emph{Selected contributions in data analysis and classification},
  pp. 161--172.

\bibitem[{Bock(2008)}]{bock2008}
Bock H (2008).
\newblock \enquote{Origins and extensions of the k-means algorithm in cluster
  analysis.}
\newblock \emph{Journal Electronique Histoire des Probabilit{\'e}s et de la
  Statistique Electronic Journal for History of Probability and Statistics},
  \textbf{4}.

\bibitem[{Chang \emph{et~al.}(2010)Chang, Qiu, Zamar, Lazarus, and
  Wang}]{Chang2010}
Chang F, Qiu W, Zamar RH, Lazarus R, Wang X (2010).
\newblock \enquote{\pkg{clues}: An \proglang{R} Package for Nonparametric
  Clustering Based on Local Shrinking.}
\newblock \emph{Journal of Statistical Software}, \textbf{33}(4), 1--16.
\newblock ISSN 1548-7660.
\newblock \urlprefix\url{http://www.jstatsoft.org/v33/i04}.

\bibitem[{Chavent \emph{et~al.}(2012)Chavent, Kuentz-Simonet, Liquet, and
  Saracco}]{Chavent2012}
Chavent M, Kuentz-Simonet V, Liquet B, Saracco J (2012).
\newblock \enquote{\pkg{ClustOfVar}: An \proglang{R} Package for the Clustering
  of Variables.}
\newblock \emph{Journal of Statistical Software}, \textbf{50}(13), 1--16.
\newblock ISSN 1548-7660.
\newblock \urlprefix\url{http://www.jstatsoft.org/v50/i13}.

\bibitem[{Cover \emph{et~al.}(1991)Cover, Thomas, Wiley \emph{et~al.}}]{Co-Th}
Cover T, Thomas J, Wiley J, \emph{et~al.} (1991).
\newblock \emph{Elements of information theory}, volume~6.
\newblock Wiley Online Library.

\bibitem[{Estivill-Castro and Yang(2000)}]{estivill2000fast}
Estivill-Castro V, Yang J (2000).
\newblock \enquote{{Fast and robust general purpose clustering algorithms}.}
\newblock \emph{PRICAI 2000 Topics in Artificial Intelligence}, pp. 208--218.

\bibitem[{Figueiredo and Jain(2002)}]{figueiredo2002unsupervised}
Figueiredo MAT, Jain AK (2002).
\newblock \enquote{Unsupervised learning of finite mixture models.}
\newblock \emph{Pattern Analysis and Machine Intelligence, IEEE Transactions
  on}, \textbf{24}(3), 381--396.

\bibitem[{Fisher(1936)}]{fisher1936use}
Fisher RA (1936).
\newblock \enquote{The use of multiple measurements in taxonomic problems.}
\newblock \emph{Annals of eugenics}, \textbf{7}(2), 179--188.

\bibitem[{Forina \emph{et~al.}(1986)Forina, Armanino, Castino, and
  Ubigli}]{forina1986multivariate}
Forina M, Armanino C, Castino M, Ubigli M (1986).
\newblock \enquote{Multivariate data analysis as a discriminating method of the
  origin of wines.}
\newblock \emph{Vitis}, \textbf{25}(3), 189--201.

\bibitem[{Fraley and Raftery(1999)}]{fraley1999mclust}
Fraley C, Raftery AE (1999).
\newblock \enquote{\pkg{MCLUST}: Software for model-based cluster analysis.}
\newblock \emph{Journal of Classification}, \textbf{16}(2), 297--306.

\bibitem[{Goldberger and Roweis(2004)}]{goldberger2004hierarchical}
Goldberger J, Roweis ST (2004).
\newblock \enquote{Hierarchical clustering of a mixture model.}
\newblock In \emph{Advances in Neural Information Processing Systems}, pp.
  505--512.

\bibitem[{Gr{\"u}nwald(2007)}]{MDLP}
Gr{\"u}nwald P (2007).
\newblock \emph{The minimum description length principle}.
\newblock MIT Press.

\bibitem[{Gr{\"u}nwald \emph{et~al.}(2005)Gr{\"u}nwald, Myung, and Pitt}]{Gr}
Gr{\"u}nwald P, Myung I, Pitt M (2005).
\newblock \emph{{Advances in minimum description length: Theory and
  applications}}.
\newblock MIT Press.

\bibitem[{Hartigan(1975)}]{Clu}
Hartigan J (1975).
\newblock \emph{Clustering algorithms}.
\newblock John Wiley \& Sons.

\bibitem[{Huang(1998)}]{huang1998extensions}
Huang Z (1998).
\newblock \enquote{Extensions to the k-means algorithm for clustering large
  data sets with categorical values.}
\newblock \emph{Data Mining and Knowledge Discovery}, \textbf{2}(3), 283--304.

\bibitem[{Jain(2010)}]{jain2010}
Jain A (2010).
\newblock \enquote{{Data clustering: 50 years beyond K-means}.}
\newblock \emph{Pattern Recognition Letters}, \textbf{31}(8), 651--666.

\bibitem[{Jain and Dubes(1988)}]{Dubes}
Jain A, Dubes R (1988).
\newblock \emph{{\it Algorithms for clustering data}}.
\newblock Prentice Hall.

\bibitem[{Jain \emph{et~al.}(1999)Jain, Murty, and Flynn}]{jain1999}
Jain A, Murty M, Flynn P (1999).
\newblock \enquote{{Data clustering: A Review}.}
\newblock \emph{ACM Computing Surveys}, \textbf{31}(3), 264--323.

\bibitem[{Kamieniecki and Spurek(2014)}]{CEC}
Kamieniecki K, Spurek P (2014).
\newblock \emph{\proglang{R} Package \pkg{CEC}}.
\newblock
  \urlprefix\url{http://cran.r-project.org/web/packages/CEC/index.html}.

\bibitem[{Korze\mbox{\'n} \emph{et~al.}(2013)Korze\mbox{\'n}, Jaroszewicz, and
  Kl\mbox{\c{e}}sk}]{korzen2013logistic}
Korze\mbox{\'n} M, Jaroszewicz S, Kl\mbox{\c{e}}sk P (2013).
\newblock \enquote{Logistic regression with weight grouping priors.}
\newblock \emph{Computational Statistics \& Data Analysis}.

\bibitem[{Kulis and Jordan(2012)}]{kulis2012revisiting}
Kulis B, Jordan MI (2012).
\newblock \enquote{Revisiting k-means: New algorithms via Bayesian
  nonparametrics.}
\newblock In \emph{Proceedings of the 29th International Conference on Machine
  Learning (ICML), Edinburgh, UK, 2012}, pp. 513--520.

\bibitem[{Kullback(1997)}]{Ku}
Kullback S (1997).
\newblock \emph{Information theory and statistics}.
\newblock Dover Pubns.

\bibitem[{Kurihara and Welling(2009)}]{kurihara2009bayesian}
Kurihara K, Welling M (2009).
\newblock \enquote{Bayesian K-means as a maximization-expectation algorithm.}
\newblock \emph{Neural computation}, \textbf{21}(4), 1145--1172.

\bibitem[{Ma \emph{et~al.}(2007)Ma, Derksen, Hong, and Wright}]{Yi_Ma}
Ma Y, Derksen H, Hong W, Wright J (2007).
\newblock \enquote{Segmentation of multivariate mixed data via lossy data
  coding and compression.}
\newblock \emph{Pattern Analysis and Machine Intelligence, IEEE Transactions
  on}, \textbf{29}(9), 1546--1562.

\bibitem[{MacKay(2003)}]{Ka}
MacKay D (2003).
\newblock \emph{Information theory, inference, and learning algorithms}.
\newblock Cambridge University Press.

\bibitem[{Massa \emph{et~al.}(1999)Massa, Paolucci, and
  Puliafito}]{massa1999new}
Massa S, Paolucci M, Puliafito P (1999).
\newblock \enquote{{A new modeling technique based on Markov chains to mine
  behavioral patterns in event based time series}.}
\newblock \emph{DataWarehousing and Knowledge Discovery}, pp. 802--802.

\bibitem[{McLachlan and Krishnan(1997)}]{EM2}
McLachlan G, Krishnan T (1997).
\newblock \emph{{The EM algorithm and extensions}}, volume 274.
\newblock John Wiley \& Sons.

\bibitem[{McLachlan and Krishnan(2007)}]{mclachlan2007algorithm}
McLachlan G, Krishnan T (2007).
\newblock \emph{The EM algorithm and extensions}, volume 382.
\newblock John Wiley \& Sons.

\bibitem[{McLachlan and Peel(2004)}]{mclachlan2004finite}
McLachlan G, Peel D (2004).
\newblock \emph{Finite mixture models}.
\newblock John Wiley \& Sons.

\bibitem[{Mirkin(2011)}]{mirkin2011choosing}
Mirkin B (2011).
\newblock \enquote{Choosing the number of clusters.}
\newblock \emph{Wiley Interdisciplinary Reviews: Data Mining and Knowledge
  Discovery}, \textbf{1}(3), 252--260.

\bibitem[{Povinelli \emph{et~al.}(2004)Povinelli, Johnson, Lindgren, and
  Ye}]{povinelli2004time}
Povinelli RJ, Johnson MT, Lindgren AC, Ye J (2004).
\newblock \enquote{Time series classification using Gaussian mixture models of
  reconstructed phase spaces.}
\newblock \emph{Knowledge and Data Engineering, IEEE Transactions on},
  \textbf{16}(6), 779--783.

\bibitem[{Sam{\'e} \emph{et~al.}(2007)Sam{\'e}, Ambroise, and Govaert}]{EM3}
Sam{\'e} A, Ambroise C, Govaert G (2007).
\newblock \enquote{{An online classification EM algorithm based on the mixture
  model}.}
\newblock \emph{Statistics and Computing}, \textbf{17}(3), 209--218.

\bibitem[{Samuelsson(2004)}]{samuelsson2004waveform}
Samuelsson J (2004).
\newblock \enquote{Waveform quantization of speech using Gaussian mixture
  models.}
\newblock In \emph{Acoustics, Speech, and Signal Processing, 2004.
  Proceedings.(ICASSP'04). IEEE International Conference on}, volume~1, pp.
  I--165. IEEE.

\bibitem[{Shannon(2001)}]{Sh}
Shannon C (2001).
\newblock \enquote{A mathematical theory of communication.}
\newblock \emph{ACM SIGMOBILE Mobile Computing and Communications Review},
  \textbf{5}(1), 3--55.

\bibitem[{Spurek \emph{et~al.}(2013)Spurek, Tabor, and Zaj{\c{a}}c}]{Sp-Ta}
Spurek P, Tabor J, Zaj{\c{a}}c E (2013).
\newblock \enquote{Detection of Disk-Like Particles in Electron Microscopy
  Images.}
\newblock In \emph{Proceedings of the 8th International Conference on Computer
  Recognition Systems CORES 2013}, pp. 411--417. Springer-Verlag.

\bibitem[{Tabor and Misztal(2013)}]{Ta-Mi}
Tabor J, Misztal K (2013).
\newblock \enquote{Detection of elliptical shapes via cross-entropy
  clustering.}
\newblock In \emph{Pattern Recognition and Image Analysis}, pp. 656--663.
  Springer-Verlag.

\bibitem[{Tabor and Spurek(2014)}]{tabor2014cross}
Tabor J, Spurek P (2014).
\newblock \enquote{Cross-entropy clustering.}
\newblock \emph{Pattern Recognition}, \textbf{47}(9), 3046--3059.

\bibitem[{Tibshirani \emph{et~al.}(2001)Tibshirani, Walther, and Hastie}]{gap}
Tibshirani R, Walther G, Hastie T (2001).
\newblock \enquote{Estimating the number of clusters in a data set via the gap
  statistic.}
\newblock \emph{Journal of the Royal Statistical Society B}, \textbf{63}(2),
  411--423.

\bibitem[{Valente and Wellekens(2004)}]{valente2004variational}
Valente F, Wellekens C (2004).
\newblock \enquote{Variational Bayesian feature selection for Gaussian mixture
  models.}
\newblock In \emph{Acoustics, Speech, and Signal Processing, 2004.
  Proceedings.(ICASSP'04). IEEE International Conference on}, volume~1, pp.
  I--513. IEEE.

\bibitem[{Wallace and Kanade(1990)}]{wallace1990finding}
Wallace RS, Kanade T (1990).
\newblock \enquote{Finding natural clusters having minimum description length.}
\newblock In \emph{Pattern Recognition, 1990. Proceedings., 10th International
  Conference on}, volume~1, pp. 438--442. IEEE.

\bibitem[{Xiong \emph{et~al.}(2002)Xiong, Chen, Wang, and
  Huang}]{xiong2002improved}
Xiong Z, Chen Y, Wang R, Huang TS (2002).
\newblock \enquote{Improved information maximization based face and facial
  feature detection from real-time video and application in a multi-modal
  person identification system.}
\newblock In \emph{Proceedings of the 4th IEEE International Conference on
  Multimodal Interfaces}, p. 511. IEEE Computer Society.

\bibitem[{Xu and Wunsch(2009)}]{xu2009clustering}
Xu R, Wunsch D (2009).
\newblock \emph{Clustering}.
\newblock Wiley-IEEE Press.

\bibitem[{Yang \emph{et~al.}(2008)Yang, Wright, Ma, and Sastry}]{Yi_Ma2}
Yang A, Wright J, Ma Y, Sastry S (2008).
\newblock \enquote{Unsupervised segmentation of natural images via lossy data
  compression.}
\newblock \emph{Computer Vision and Image Understanding}, \textbf{110}(2),
  212--225.

\bibitem[{Zhang \emph{et~al.}(2004)Zhang, Zhang, and Yi}]{zhang2004competitive}
Zhang B, Zhang C, Yi X (2004).
\newblock \enquote{Competitive EM algorithm for finite mixture models.}
\newblock \emph{Pattern recognition}, \textbf{37}(1), 131--144.

\end{thebibliography}
\end{document}